%% file: main_file.tex
\documentclass[sigconf]{acmart}

\AtBeginDocument{%
  \providecommand\BibTeX{{%
    \normalfont B\kern-0.5em{\scshape i\kern-0.25em b}\kern-0.8em\TeX}}}

\setcopyright{acmlicensed}
\copyrightyear{2018}
\acmYear{2018}
\acmDOI{XXXXXXX.XXXXXXX}

\acmConference[KDD'24]{}{August 25--29,
  2024}{Barcelona, Spain}
\acmISBN{978-1-4503-XXXX-X/18/06}

\usepackage{makecell}
\usepackage{multirow}
\usepackage{algorithm}
\usepackage{algpseudocode}
\usepackage{tabularx}
\usepackage{graphicx}
\usepackage{enumitem}
\usepackage{multicol}
\newcommand{\systemname}{\textit{LiGNN}}
\newcounter{savealgorithm}

\makeatother
\algnewcommand\algorithmicforeach{\textbf{for each}}
\algdef{S}[FOR]{ForEach}[1]{\algorithmicforeach\ #1\ \algorithmicdo}

\algdef{SE}[REPEATN]{RepeatN}{End}[1]{\algorithmicrepeat\ #1 \textbf{times}}{\algorithmicend}

\begin{document}

\title{LiGNN: Graph Neural Networks at LinkedIn}


\author{Fedor Borisyuk, Shihai He, Yunbo Ouyang, Morteza Ramezani, Peng Du, Xiaochen Hou,\\Chengming Jiang, Nitin Pasumarthy, Priya Bannur, Birjodh Tiwana, Ping Liu, Siddharth Dangi,\\ Daqi Sun, Zhoutao Pei, Xiao Shi, Sirou Zhu, Qianqi Shen, Kuang-Hsuan Lee, David Stein$^{*}$, Baolei Li$^{*}$, Haichao Wei, Amol Ghoting, Souvik Ghosh}
\affiliation{\institution{LinkedIn Inc.}}

\newcommand{\etal}{\textit{et~al.}}
\renewcommand{\shortauthors}{Fedor Borisyuk, \etal}

\begin{abstract}
\input{abstract}

\end{abstract}



\maketitle

\def\thefootnote{*}\footnotetext{Work done while at LinkedIn.}
\input{intro.tex}

\input{related_work.tex}
\input{overview.tex}
\input{Training_stability_speed.tex}

\input{nearline_inference.tex}

\input{experiments.tex}

\section{Deployment Lessons}\label{sec:deployment_lessons}
\input{deployment_lessons.tex}
\section{Conclusion}\label{sec:conclusion}
\input{conclusion.tex}


\bibliographystyle{ACM-Reference-Format}
\bibliography{bibliography}

\include{appendix}

\end{document}

%% file: abstract.tex
In  this  paper, we  present {\systemname}, a deployed large-scale Graph Neural Networks (GNNs) Framework. We share our insight on developing and deployment of GNNs at large scale at LinkedIn. We present a set of algorithmic improvements to the quality of GNN representation learning including temporal graph architectures with long term losses, effective cold start solutions via graph densification, ID embeddings and multi-hop neighbor sampling. We explain how we built and sped up by 7x our large-scale training on LinkedIn graphs with adaptive sampling of neighbors, grouping and slicing of training data batches, specialized shared-memory queue and local gradient optimization. We summarize our deployment lessons and learnings gathered from  A/B test experiments. 
The techniques presented in this work have contributed to an approximate relative improvements of 1\% of Job application hearing back rate, 2\%  Ads CTR lift, 0.5\% of Feed engaged daily active users, 0.2\% session lift and 0.1\% weekly active user lift from people recommendation. 
We believe that this work can provide practical solutions and insights for engineers who are interested in applying Graph neural networks at large scale.

%% file: intro.tex
\section{Introduction}\label{sec:intro}
\begin{figure}[t]
    \centering
    \includegraphics[height=7cm,width=9cm]{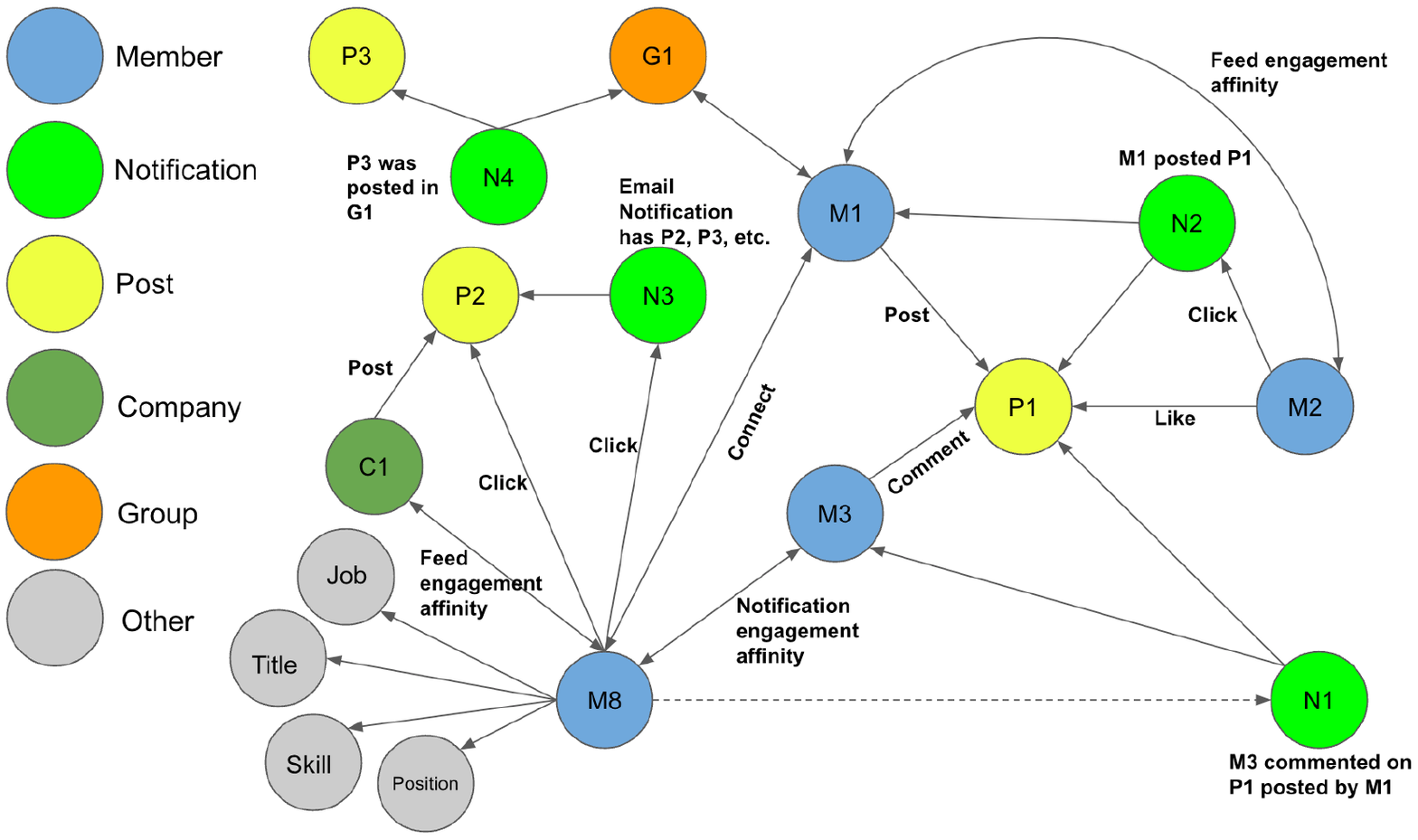}
    \caption{Schematic representation of LinkedIn Graph. Members engaging with Posts, Jobs, Groups, Companies and other members.}
    \label{fig:LinkedInGraph}
    \vspace{-2.0em}
\end{figure}

LinkedIn is the world's largest professional network with more than 1 billion members in more than 200 countries and territories worldwide. LinkedIn's ecosystem encompasses members, companies, universities, students, and groups, all forming connections within the professional graph. Hundreds of millions of LinkedIn members actively engage to seek opportunities and connect with professionals. Integrating all these entities into the graph is an intuitive step. Our graph boasts up to a hundred billion nodes and several hundred billion edges. Graph edges symbolize various activities on the LinkedIn app, such as job applications, post engagements, and networking interactions (see Figure \ref{fig:LinkedInGraph}).

Developing {\systemname} at scale presented several challenges:
\begin{itemize}[leftmargin=*]
\item GNN training at scale: Unlike traditional DNN training, GNN training has unique training scalability issues due to graph hosting requirements (\S\ref{sec:Overview:TrainingInfra}). 
\item Diverse entities: our goal was to create a unified graph embedding space for various entities like posts, members, companies, and jobs (\S\ref{sec:Overview:GraphConstruction}). 
\item Cold start: infrequent visits by some LinkedIn members result in limited preference data (\S\ref{sec:coldstart}). 
\item Dynamic system: the dynamic, temporal nature of the LinkedIn ecosystem limited the capabilities of GNN models.
\end{itemize}


Our technological advancements in Graph Neural Networks address these challenges. The \textbf{contribution of the paper} consists of:
\begin{itemize}[leftmargin=*]
     \item GNN training at scale: section \S\ref{sec:Overview:TrainingInfra} outlines the GNN training infrastructure, while \S\ref{sec:training_stability_speed} discusses scaling LinkedIn's GNN training. This involves integrating Microsoft's real-time DeepGNN graph engine ~\cite{deepGNNref} with GPU node training jobs. We propose novel techniques using adaptive sampling and node grouping to expedite training, and share our approaches on effective data processing with shared memory queue. We also share our path on how we achieved high GNN training stability (\S\ref{sec:stability}).
     \item Diverse entities: in \S\ref{sec:Overview:GraphConstruction} we share our experience of building and optimizing a graph for multi-task environments at LinkedIn scale, integrating different LinkedIn entities into a singular embedding space.
     \item Cold start: to improve the experience for less active members, we propose methods for graph densification (\S\ref{sec:coldstart}) and share our experience on speeding up multi-hop graph sampling (\S\ref{sec:multihop}). 
     \item Dynamic system: to keep up with LinkedIn's dynamic ecosystem, we integrated Temporal Graphs with long-term optimization (\S\ref{sec:temporal_graphs}) and implemented near-line graph serving (\S\ref{sec:nearline_inference}). Temporal modeling helps GNNs discern historically significant member interactions, while the near-line graph infrastructure swiftly reflects interactions in member and item representations. Our solution to incorporate temporal aspects in graphs is simpler and more scalable, making it suitable for production, compared to conventional temporal graph architectures \cite{rossi2020temporal, skarding2021foundations}.  
\end{itemize}

%% file: related_work.tex
\section{Related Work}\label{sec:related_work}
Graph Neural Networks (GNNs) are effective for modeling graphs \cite{GraphSage_paper} and relational data~\cite{fey2023relational}. Much research has focused on enhancing GNN model architectures~\cite{GraphSage_paper, rossi2020temporal, GINpaper, RGCN_paper}. Our work builds upon the SAGE~\cite{GraphSage_paper} architecture, integrating sequential temporal modeling with transformer-based sequence modeling and long-term losses~\cite{pancha2022pinnerformer, rangadurai2022nxtpost}, tailored to the GNN domain.

GNNs propagate signals across graphs, where neighbor sampling is crucial. We found Personalized PageRank (PPR)~\cite{bojchevski2020scaling, GraphSage_paper, FORA_paper, TopPPR_paper} sampling to be more effective than random sampling. Our implementation uses two-hop sampling with forward push and random walks~\cite{bojchevski2020scaling}, showing scalability on LinkedIn data.

Several studies have aimed to accelerate GNN training jobs at industry scale such as MLPinit~\cite{han2023mlpinit}, GraphStorm~\cite{GraphStorm_paper}, BigGraph~\cite{lerer2019pytorchbiggraph}, HUGE~\cite{HUGE_paper}. We introduce over three novel techniques in \S\ref{sec:training_stability_speed} that achieved a significant reduction in training time on large-scale production data.

GNNs are widely used in the industry for various applications like people recommendations~\cite{Friend_GNN_snap_paper}, Anti-abuse ~\cite{Grale_paper}, weather forecasting ~\cite{lam2022graphcast}, Ads \cite{Ads_GNN_paper}. Our system, evaluated within LinkedIn, shows promising results in domains of People recommendations, Ads, Job Recommendations, and Feed post recommendations.

Numerous studies have explored cold start solutions, such as student-teacher consistency learning~\cite{Consistency_gnn_paper}, attribute edges~\cite{Qian_2022}, self-supervised pre-training \cite{hao2020pretraining}, meta-learning ~\cite{hao2020pretraining, Multi_Strategy_Based_paper}, and similarity-based embeddings with LSH bucketing \cite{Grale_paper}. Our approach introduces artificial nearest neighbor edges to cold start nodes, leveraging content embeddings, which has shown quality improvements in various production applications at LinkedIn.

%% file: overview.tex
\section{GNN Modeling and Training}\label{sec:modeling}

\input{Graph_construction.tex}

\input{TrainingInfra.tex}

\input{gnn_architecture.tex}

\subsection{Temporal Graphs}\label{sec:temporal_graphs}
\input{Temporal_graphs}

\input{coldstart.tex}

\input{multihop.tex}

\input{llm_gnn.tex}

%% file: Graph_construction.tex
\subsection{Graph Construction}\label{sec:Overview:GraphConstruction}
The graph used for the LinkedIn GNN models is a heterogeneous graph, which contains tens of node types and edge types, as shown in Figure~\ref{fig:LinkedInGraph}. To densify the graph, we combine the subgraphs from different domains together, such as feed recommendations, job recommendions, notifications. Each domain can train their GNN models using its owned subgraph, or leveraging the combined graph. Over all, the graph contains 3 types of edges: (1) engagement edges, (2) affinity edges and (3) attribute edges. The engagement edges represent the engagements between LinkedIn's members and the contents on the LinkedIn platform, such as "member M2 liked post P1" is represented by an edge between M2 and P1. The affinity edges capture the historical engagements between LinkedIn's members and the creator of the contents, such as "member M2 has engaged with contents posted on LinkedIn by member M1" is represented by an edge between M2 and M1. The attribute edges capture the HAS-A relationships between two nodes such as "member M8 has a software engineer job" is represented by an edge between M8 and the corresponding job node. The edges are weighted by the strength of the affinity or engagements between two entities, except all the attribute edges use 1.0 as their edge weights. Currently, the combined graph contains up to a hundred billion nodes and several hundred billion edges.

%% file: TrainingInfra.tex
\subsection{Training infra}\label{sec:Overview:TrainingInfra}
\begin{figure}[h]
  \centering
  \includegraphics[width=1.0\columnwidth]{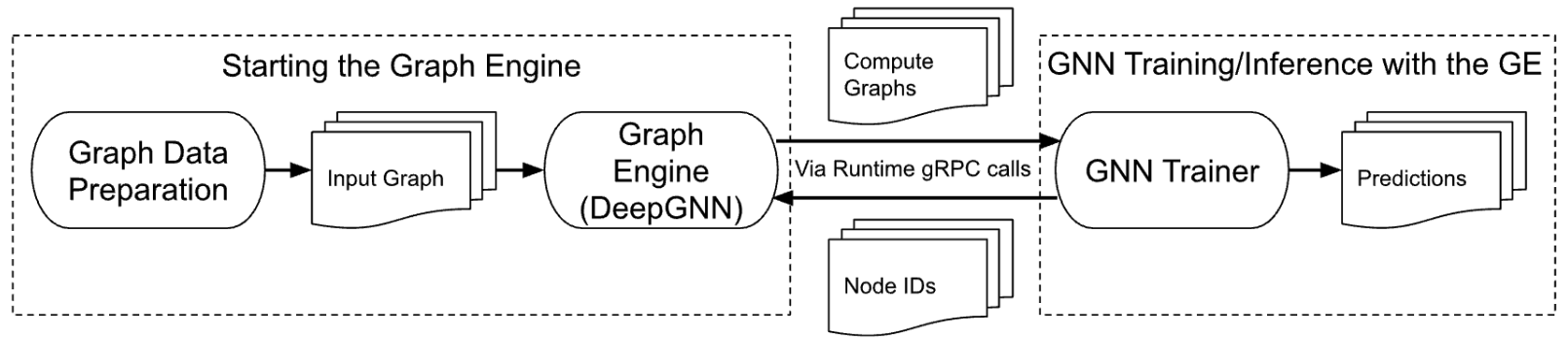}
  \caption{High level view of GNN pipelines.}
  \label{fig:gnnPipeline}
  \vspace{-1.0em}
\end{figure}

In LinkedIn, all GNN training and inference jobs are executed in the in-house Kubernetes (K8S)~\cite{kubernetes} based cluster, which has access to a Hadoop File System (HDFS). 
Each job needs to deploy a Graph Engine (GE, usually on CPU nodes) and a GNN Trainer (usually on GPU nodes) in the K8S cluster. The lightweight high-performance Microsoft DeepGNN~\cite{deepGNNref} was chosen as the Graph Engine to provide fast real-time graph sampling with a variety of sampling strategies. As shown in Figure~\ref{fig:gnnPipeline}, the Graph Data Preparation step collects the graph data including edges and node features, and writes the data to HDFS. During training or inference, the GE loads the graph data into distributed memory and serves the data in real-time. Then, the GNN Trainer makes gRPC calls to the GE to fetch the sampled compute graphs and use the data for GNN training or inference. The GE can be deployed independently or as a part of specific training or inference job. Depending on the size of graph, one can launch one or more instances (pod\footnote{Pods are the smallest deployable units of computing in Kubernetes.}) to serve a portion of the partitioned graph. During training (or inference) the DeepGNN client queries the GEs with a given setup, which consists of the sampling algorithm and configuration, over \href{https://grpc.io/}{gRPC}. The resulting data is consumed by the underlying deep learning framework (Tensorflow).
Figure~\ref{fig:gnnPipeline} summarizes the GNN pipelines on K8S cluster at Linkedin.


%% file: gnn_architecture.tex
\subsection{GNN architecture}\label{sec:gnn_architecture}
\begin{figure}[h]
  \centering  \includegraphics[width=0.8\columnwidth]{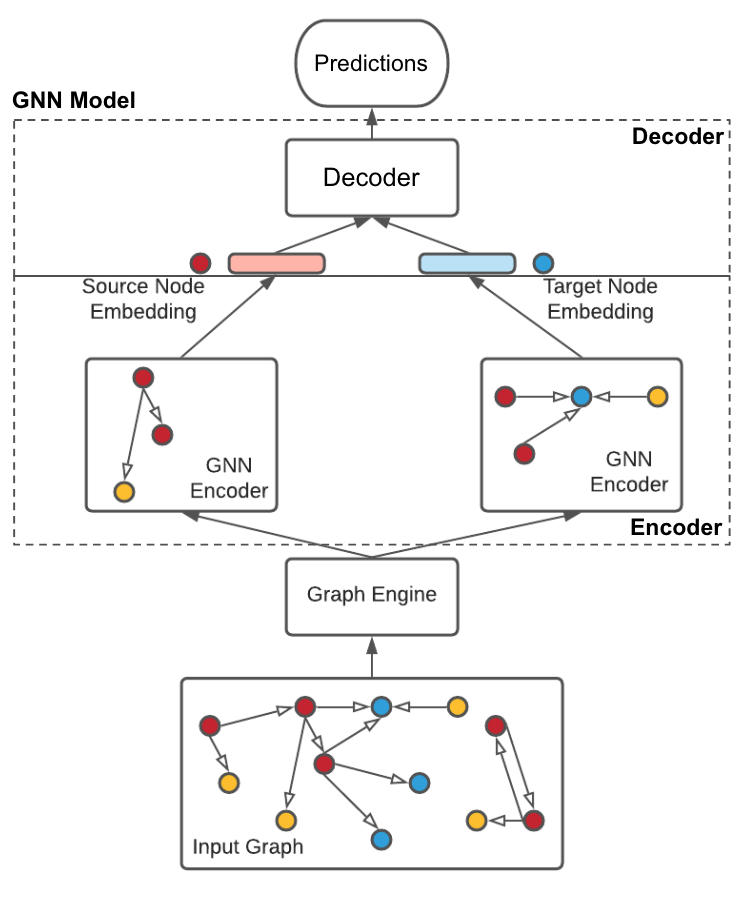}
  \caption{GNN model architecture.}
  \label{fig:gnnModel}
  \vspace{-2.0em}
\end{figure}

Considering the complexity of using GNN models to replace the existing machine learning models in LinkedIn, we adopted the the encoder-decoder architecture for the GNN models as shown in Figure ~\ref{fig:gnnModel}. In this way, we can only take the trained encoder to generate the node embeddings and apply the embeddings in the downstream application models as new features. To handle the large scale LinkedIn graph and carry out inductive learning, the encoder adopts the GraphSAGE-style framework ~\cite{GraphSage_paper}, which inductively generates the node embeddings based on graph sampling and neighborhood aggregation. The graph sampling is provided by the DeepGNN GE, including multi-hop random sampling, weighted sampling, Personalized PageRank (PPR) sampling. The neighborhood aggregation mainly uses the mean aggregation or the attention-based aggregation. The decoder of the GNN model takes the embeddings generated from the encoder as its input and computes the predictions. Currently we support Multilayer Perceptron (MLP) decoder, cosine decoder and in-batch negative sampling decoder \cite{que2search_paper} for link prediction tasks. The cosine decoder computes the cosine similarity between the source node embedding and the destination node embedding, and makes predictions on it. The in-batch negative sampling decoder treat all other samples in the batch as negative samples and makes predictions based on the dot products of each sample pairs. In the next section we discuss how we add transformer based architectures to {\systemname}.

%% file: Temporal_graphs.tex
\begin{figure*}
    \centering
    \includegraphics[width=\textwidth]{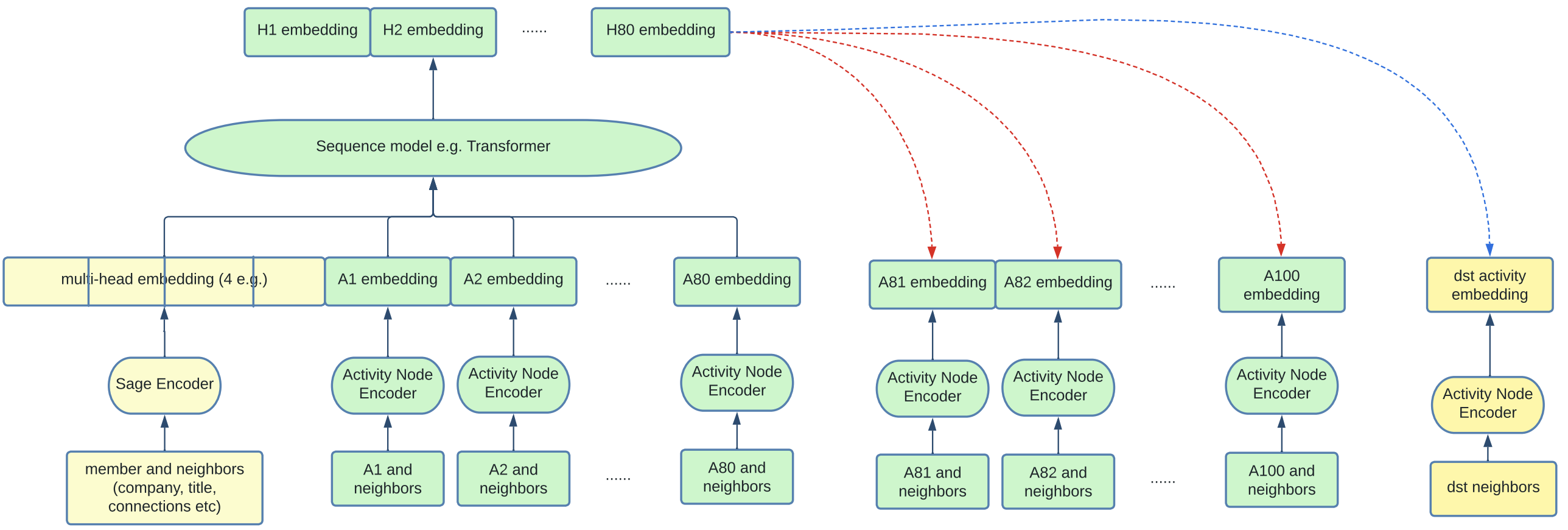}
    \caption{Temporal model includes: (1) static SAGE-encoder depicted in Yellow, (2) transformer based temporal sequence model in Green. Red lines depict long term loss and blue lines depict cosine similarity loss described in \S\ref{sec:gnn_architecture}.}
    \label{fig:temporal_model}
\end{figure*}
\begin{table}[ht]
\centering
\small 
\begin{tabular}{l|c|c}
\hline
\textbf{Experiment set up} & AUC & Relative Lift \\ \hline
Baseline: SAGE encoder + BCE loss &  $0.71978$ & -- \\ \hline
\makecell{Temporal Encoder (TempEnc)} &$0.75204$ & 4.48\% \\ \hline
\makecell{TempEnc + Positional encoding} &$0.75277$ & 4.58\% \\ \hline
\makecell{TempEnc + timestamp encoding} &$0.75143$ & 4.40\% \\ \hline
\makecell{TempEnc + Regular Causal mask} &$0.74991$ & 4.19\% \\ \hline
\makecell{TempEnc + Prefix Causal Mask} &$0.75316$ & 4.64\% \\ \hline
\makecell{TempEnc + use dst neighbors in src} &$0.75978$ & 5.56\% \\ \hline
\makecell{TempEnc + long term loss, future history len = 40 }&$0.75157$ & 4.42\% \\ \hline
\makecell{TempEnc + long term loss, future history len = 20 }&$0.75099$ & 4.34\% \\ \hline
\makecell{TempEnc + long term loss, future history len = 10 } &$ 0.75193$ & 4.47\% \\ \hline
\makecell{all above combined (future history length = 10) } &$0.76176$ & 5.83\% \\ \hline

\end{tabular}
\caption{Temporal model experiment results on Feed data.}
\label{table:jobs_recommendations_table}
\vspace{-2.0em}
\end{table}


The GNNs that we discussed so far are static, which lack temporal dynamics that is critical for professional social networks like LinkedIn, where interactions are time-sensitive. The rise of transformer models in recommendation systems, particularly for sequential data, suggests a natural fusion of GNNs with temporal sequence modeling~\cite{pancha2022pinnerformer, rangadurai2022nxtpost, awad2023adsformers}. Although there is research on event-driven and message-passing in dynamic GNNs~\cite{rossi2020temporal, skarding2021foundations}, their real-world applicability is limited. We redefine "temporal graphs" to focus on temporal sequence modeling within GNNs, as shown in Figure ~\ref{fig:temporal_model}.

Our design modifies the neighbor sampling method of a SAGE~\cite{ying2018graph} encoder by adding time-based node sampling to capture the last N (e.g. 100) activities of member before a certain time. We expand the SAGE encoder's output to multi-head dimensions with head number $H$ (e.g. 4) by using a larger dimension equals $H * d$, reshaping it into a sequence with length $H$ and dimension $d$. After that we encode the $N$ activities with a node encoding module and concatenate them with the outputs of the SAGE encoder and eventually get a sequence of length $H + N$ and dimension $d$. Then we feed this sequence to the transformer's encoder to get an output of length $H + N$ and dimension $d$. We also add positional encoding ~\cite{vaswani2017attention} and prefix causal masking where for the first $H$ tokens, we have full attention and for each token in the last $N$ positions, they can only attend all the $H$ tokens and tokens before itself in the sub-sequence of the $N$ activity embeddings~\cite{liu2018generating}. We combined binary cross entropy loss and a long term losses ~\cite{rangadurai2022nxtpost, pancha2022pinnerformer} during training. We can cut the the length-$N$ sequence into two parts: first part length equal $N_1$ and second part length equal $N_2$, where $N_1 + N_2 = N$. Long Term Loss extends prediction to $N_2$ future events, where we can use the output embedding at positions $N_1$ to predict embeddings from $N_1$ to $N$.

%% file: coldstart.tex
\subsection{Graph Densification}\label{sec:coldstart}

The degree distribution in social network graphs often follows a power law, with most nodes having few interactions. This presents a challenge for neighborhood aggregation in GNNs, particularly for nodes with low out-degrees. To combat this, {\systemname} implements graph densification by adding artificial edges based on auxiliary information. When a low-out-degree node is similar to a high-out-degree node, an artificial edge is added between them, leveraging external content embeddings to gauge the similarity. The graph densification algorithm is outlined in Algorithm \ref{algo:cs}. 

Algorithm \ref{algo:cs} consists of three main functions. The Query function retrieves the embedding for a node. the approximate\_knn function identifies the top $k$ similar high-out-degree nodes for a low-out-degree node, using embedding similarity. For scalability in handling numerous nodes, we use an in-house approximate nearest neighbor search solution, based on HNSW \cite{doshi2020lanns}. The create\_edge function forms artificial edges between a low-out-degree node and its top $k$ similar high-out-degree counterparts. This method facilitates information flow from active nodes to less active nodes, mitigating cold start issues.


In LinkedIn's production scenarios, based on node out-degree quantiles we set degree\_upper\_bound at the 90th percentile, and degree\_lower\_bound set at the 30th percentile, with optimal results when $k$ is around 50. We utilize different external embeddings for varying node types, such as profile LLM embeddings for member nodes, derived from member profile data, and content embeddings for item nodes, based on text and image content.


\begin{algorithm}
\small
\caption{Graph Densification}\label{algo:cs}
\begin{algorithmic}[1]
    \Require degree\_lower\_bound, degree\_upper\_bound, external\_embeddings, $k$
    
    \Function{Graph\_Densification}{degree\_lower\_bound, degree\_upper\_bound, external\_embedding, $k$}
        \State low\_degree\_node\_set $\gets$ \{$x$|$x$'s out-degree $\leq$ degree\_lower\_bound \}
        \State high\_degree\_node\_set $\gets$ \{$x$|$x$'s out-degree $\geq$ degree\_upper\_bound \}
        \State high\_degree\_embedding\_set $\gets$ \{query(external\_embeddings, node) for node in high\_degree\_node\_set\}        
        \For{node \textbf{in} low\_degree\_node\_set}
            \State node\_embedding $\gets$ query(external\_embeddings, node)
            \State top\_k\_set $\gets$ approximate\_knn(high\_degree\_embedding\_set, $k$, node\_embedding)
            \State create\_edge(top\_k\_set, node)   
        \EndFor
    \EndFunction
\end{algorithmic}
\end{algorithm}

%% file: multihop.tex
\subsection{Multi-hop Graph Sampling}\label{sec:multihop}
For {\systemname} to surpass traditional deep learning methods, effective sampling is key. Simple one-hop sampling falls short in capturing the complex graph topology, hence {\systemname} adopts multi-hop sampling for deeper graph analysis. Three multi-hop sampling techniques were explored:
\begin{itemize}
\item \textbf{Multi-hop random/weighted sampling}: This method allows for either random sampling or user-configurable weighted sampling, where weights are adjustable for different edge types.
\item \textbf{Multi-hop Personalized PageRank (PPR) Sampling}: Integral to {\systemname}, PPR is a prominent tool in large-scale graph mining. It locates neighbors with the top $k$ PPR scores relative to a source node, identifying key topological nodes. Despite PPR's slower pace compared to random/weighted sampling, efficiency is enhanced through approximate calculations using the Forward Push Algorithm and system-level optimizations. To accelerate PPR sampling for a batch of nodes, we consolidate sampling requests in each iteration of Forward Push into a single batch, reducing overhead.
\item \textbf{Two-hop Personalized PageRank (PPR) Sampling}: Tailored for nearline serving, which currently only supports 2-hop methods, this approach returns neighbors within a 2-hop radius with the top $k$ PPR scores. It utilizes a fast 2-hop random walk algorithm for PPR computation, offering quicker sampling than multi-hop PPR.
\end{itemize}
Subsequent sections will demonstrate the superiority of PPR sampling over multi-hop random/weighted sampling in terms of effectiveness. In experiments in Follow Feed and People recommendations, 2-hop PPR sampling contributes around 90\% of gains and accelerate the sampling speed by 3 times, therefore we choose 2-hop PPR sampling as the default sampling strategy.

%% file: Training_stability_speed.tex
\section{Training Stability and Speed}\label{sec:training_stability_speed}
In this section we will cover how we improved the stability and speed of GNN model training at LinkedIn.

\begin{algorithm}
\small
\caption{Adaptive\_Neighbor\_Sampling} \label{algo:adaptive_neghbor_sampling}
\begin{algorithmic}[1]
    \Require starting\_neighbor\_count, final\_neighbor\_count, metric, tolerance, stride,  tolerance\_decay, min\_update\_freq, last\_metric = 0.0
    
    \Function{Adaptive\_Neighbor\_Sampling}{starting\_neighbor\_count, final\_neighbor\_count, metric, tolerance, last\_metric, stride}
        \State current\_neighbor\_count $\gets$ starting\_neighbor\_count
        
        \For{epoch \textbf{in} epochs}
           \For{step \textbf{in} steps}
                \State train\_data $\gets$ query(current\_neighbor\_count)
                \State train(model, train\_data)
            \EndFor
            \State current\_metrics $\gets$ evaluate(model, val\_data)
            
            \If{current\_metric $\leq$ last\_metric + tolerance or epoch \% min\_update\_freq == 0}
                \State current\_neighbor\_count $\gets$ min(current\_neighbor\_count + stride, final\_neighbor\_count) 
            \EndIf
            \State current\_metric $\gets$last\_metric
            \State tolerance $\gets$ tolerance * tolerance\_decay
            
        \EndFor
        
        \State \textbf{return} model
    \EndFunction
\end{algorithmic}
\end{algorithm}
\begin{algorithm}
\small
\caption{Grouping and Slicing}\label{algo:grouping_slicing}
\begin{algorithmic}[1]
    \Require training\_data, group\_size, gradient\_step    
    \Function{Grouping\_Slicing\_Sampling}{training\_data, group\_size, gradient\_step}
        \State grouped\_training\_data $\gets$ group(training\_data, key="member\_id")
        \State sliced\_padded\_data $\gets$ slice\_and\_pad(grouped\_training\_data, group\_size)
        \Comment{pad dummy item ids and labels if count is smaller than group\_size}
        \For{epoch \textbf{in} epochs}
            \For{step \textbf{in} steps}
                \State training\_batch $\gets$ query(sliced\_padded\_data)
                \Comment{each row in training\_batch contains one member and group\_size items}
                \State local\_item\_size  $\gets$ group\_size \% gradient\_step
                \For{i \textbf{in} gradient\_step}
                    \State member\_prediction $\gets$ model(training\_batch["member"])
                    \State item\_data, mask $\gets$ training\_batch["item"][$i \times 
                    local\_item\_size$ : $(i + 1) \times local\_item\_size$]
                    \Comment{mask is to tell what items are padded}
                    \State item\_prediction $\gets$ model(item\_data)
                    \State loss $\gets$ cal\_loss(member\_prediction, item\_prediction, label, mask)
                    \State update\_model(model, loss)
                \EndFor            
            \EndFor
        \EndFor
        \State \textbf{return} model
    \EndFunction
\end{algorithmic}
\end{algorithm}

\subsection{Training stability}\label{sec:stability}
Training GNN models, unlike conventional deep learning models, demands real-time graph sampling from the GE and retrieving labeled data from HDFS, intensifying network strain and affecting training stability. We implemented several techniques, boosting the training success rate from 30\% to over 90\%, as detailed in Table \ref{table:training_stability}.

\textbf{gRPC Retry:} GNN training workers often fetch GBs of data from the GE for each batch via gRPC calls, straining the data transmission between workers and the GE server. With distributed training employing 6 to 24 workers, connection losses to the GE were common. By modifying the default gRPC retry policy to maximize "max\_attempts" and "max\_backoff", we effectively resolved the connection issue, enhancing the training success rate by 15\%.

\textbf{Horovod Training:} Besides connection problems with the GE server, many job failures stemmed from worker-to-worker communication breakdowns. Transitioning from TensorFlow's MultiWorkerMirroredStrategy to Horovod distributed training, which utilizes NVIDIA's NCCL 2 and ring allreduce operation, significantly improved training stability, increasing success rates by 35\%.

\textbf{Memory Leak:} Parallel data fetching in training workers, involving multiple prefetchers for graph data and storing batch data in a queue with a typical size of 10, usually consumed tens of GBs of memory. We observed delayed garbage collection, leading to memory leaks and out-of-memory failures. Adopting TensorFlow's GeneratorEnqueuer resolved this memory leak issue, further enhancing training stability by 10\%.

\subsection{Training speed}\label{sec:training_speed}
Increasing model size typically boosts performance, but it also lengthens training time. Thus, we focus on techniques to accelerate training, enabling swift iteration even as the model grows. We've also observed that GNN jobs are often data-bound, implying that optimizing neighbor collection from the graph engine can significantly impact training speed. Overall during development of GNNs at LinkedIn the training time reduced from 24 hours, when we started, to 3.3 hours on the latest training jobs, with largest contributions from Adaptive Neighbor sampling, Grouping and Slicing and Share-Memory Queue.

\textbf{Reduce average step time:} Typically, each training step is comprised of three components: data loading, forward and backward pass. If data-parallel distributed training is used, gradients need to be communicated across all workers through an AllReduce operation after backward pass. To reduce average step time, we can focus on optimization of the most time consuming components. Local gradient aggregation is a technique to reduce the frequency of gradient communication. Gradients will be aggregated locally on each worker for N mini-batches before they are sent to other workers through AllReduce. Note that local gradient aggregation is effectively increasing batch size by N times, and utilizing techniques like learning rate scaling ~\cite{lr_scaling_ref} is important for large-batch training.

Mixed precision training ~\cite{mixed_precision_ref} combines the use of half precision and single precision numerical formats in math operations. With GPU support, it brought 8\% speedup in forward and backward pass of our GNN model training. Numerical underflow and overflow issues can occur under float16 computation. To avoid it, operations with large reductions should be carried out in float32, model's output layer should also use float32 to guarantee accurate calculation of loss. In our experiments, keeping the last layer of node aggregator in float32 is also crucial to maintaining model accuracy under mixed precision training.

\textbf{Increase convergence speed:} We explored MLPinit \cite{han2023mlpinit}, which trains node encoder weights from the node features in two tower style link-prediction matching without querying the GE. We observed 16.25\% speedup from using MLPinit. Next we will show how we generalized MLPinit using Adaptive Neighbor Sampling strategy to decrease training speed even further.

\textbf{Adaptive Neighbor Sampling}: Since the I/O (reading data from GE) is the bottleneck of GNN training, we proposed several techniques to speeding up GNN by tackling the I/O part, one of which is to adaptively increase the number of neighbors to be sampled during training. We sample a small number of neighbors at the beginning and adaptively increase the neighbor count by monitoring the model performance. If the metric (e.g., AUC) keeps increasing with a small number of neighbors, we do not sample more neighbors. We only sample more neighbors when the metrics are not improved by a certain threshold. Since the number of neighbors and I/O time are correlated, starting with a small number of neighbors to learn a model can help save a large amount of training time (Algorithm \ref{algo:adaptive_neghbor_sampling}). 


\textbf{Grouping and Slicing:} 
The training dataset comprises millions of triplets including member IDs, item IDs (like follow feed posts or jobs), and labels, showing interactions between members and items. Notably, active members often interact with numerous items, confirmed by feed dataset analysis. Given the I/O constraints of GNN, traditional feature generation by querying neighbors for each member-item pair is inefficient due to repeated queries for active members. To optimize, we group training records by members, slicing grouped items and labels at a set threshold, then querying once for the member and grouped items. For instance, if a member has 10 interactions and the group size is 5, we create two data records for this member with 5 items each, cutting GE queries from 10 to 2, albeit each query being slightly more extensive.


Once we get the grouped data, e.g., one member with 5 items, there are two training approaches: (A) generate member and item embeddings together, compute average loss of the 5 pairs, backpropagate once, or (B) forward and backward passes for each pair, updating the model 5 times. While A is generally faster, it may underperform compared to B. However, with large model sizes, A can be a good way to reduce training time. We made the number of training passes configurable: it can be any divisible number between (A) and the group size (B). Experiments on LinkedIn data showed that using an intermediate number performs more effectively without reducing model quality and leads to a 69.9\% reduction in training time. For more details, refer to Algorithm \ref{algo:grouping_slicing} for full details. See Table \ref{table:training_time_reduction}, Figure \ref{fig:feed-adp-grp-slc} and Figure \ref{fig:lts:lts-adp-grp-slc} in appendix for results of experiments on Follow Feed and Job recommendations.


\textbf{Python Multi-Processing with Shared Memory Queue:} 
DeepGNN leverages C++ for computation-heavy tasks like sampling and querying node and edge features, but LinkedIn's models and sampling algorithms are primarily in Python. Python's operations, constrained by the ~\href{https://wiki.python.org/moin/GlobalInterpreterLock}{GIL}, involve considerable data processing and generation. To manage DeepGNN client's extensive data processing during training, we adopted Python Multi-Processing for parallel prefetching and pre-processing of data batches. However, using multiple Python processes entails overhead from data copying between parent and child processes. To minimize this, we crafted a shared-memory queue in native Python, employing the multiprocessing package to simultaneously query the DeepGNN Graph Engine across multiple processes. This approach efficiently prefetches and preprocesses the necessary training data. Our experiments demonstrated that this multi-processing with a shared-memory queue can reduce training times by as much as 68\%.

\textbf{GPU co-location:} We also explored GPU co-location used in other graph engines ~\cite{GraphStorm_paper}, where we locate Graph Engine on the CPUs of GPU machines to save some TF-to-GE communication. However we didn't observe improvement in training speed. We observed that currently our training jobs are more constrained on TF-to-TF communication in comparison to TF-to-GE communication, leading to diminishing returns from co-location. The situation can change depending on the models we develop in the future.

\begin{table}[ht]
\centering
\small 
\begin{tabular}{l|cc}
\hline
\textbf{Technique} & Train Success Improvement \\ \hline
gRPC Retry & $15\%$ \\
Horovod Training & $35\%$ \\
Data Generator with GeneratorEnqueuer & $10\%$ \\ \hline
\end{tabular}
\caption{Training Stability}
\label{table:training_stability}
\vspace{-2.0em}
\end{table}

\begin{table}[ht]
\centering
\small 
\begin{tabular}{l|c}
\hline
\textbf{Technique} & Training time reduction \\ \hline
{MLPinit} & $16.25\%$  \\ 
{Adaptive neighbor sampling} & $24.2\%$ \\
{Grouping and Slicing} & $69.9\%$ \\
Mixed Precision & $8.0\%$ \\
LGA & $35.2\%$ \\
GPU Co-location & $0.0\%$ \\
Shared-Memory Queue & $68.04\%$ \\ \hline
\end{tabular}
\caption{Training Time reduction techniques measured on one of our largest Follow Feed dataset. Training time reduced from 24 hours, when we started, to 3.3 hours.}
\label{table:training_time_reduction}
\vspace{-2.0em}
\end{table}

%% file: nearline_inference.tex
\section
{Near-line inference}\label{sec:nearline_inference}
The online production system at LinkedIn places high importance on feature freshness, particularly when members interact with posts. To avoid stale recommendations from outdated GNN embeddings, a GNN model inference pipeline is used for near real-time generation of member/item GNN embeddings. The term "Item" encompasses recommendations like Posts, Jobs, Ads, and People. 

This section covers two areas: the tech stack of the nearline pipeline and its architecture. LinkedIn's nearline pipeline utilizes \href{https://beam.apache.org/}{Apache Beam}. The Managed-beam team and the Machine Learning Infrastructure team at LinkedIn have contributed valuable components, such as SourceComponent, SinkComponent, and InferenceComponent, to assist AI engineers in minimizing development costs. However, challenges like the lack of batch feature fetchers, data converters for 2D tensors, and certain sampling functions were noted. To improve the pipeline for GNN embeddings, LinkedIn developed specific Beam components, aiming for wide applicability in GNN use cases. Integrating these with LinkedIn's ecosystem, particularly the \href{https://www.slideshare.net/DavidStein1/frame-feature-management-for-productive-machine-learning}{Frame} framework and ML infra data types, was a significant challenge.

The GNN nearline pipeline, depicted in Figure~\ref{fig:GNN_nearline}, starts with an Item Creation event via Kafka, when member interacts with an Item (e.g., clicks, connects, applies). It performs joins to collect features for the GNN model, which then conducts inference. Outputs are stored in \href{https://engineering.linkedin.com/blog/2022/open-sourcing-venice--linkedin-s-derived-data-platform} {Venice} feature storage or a Kafka topic for other Beam pipelines. This process also applies to member updates, with an added feature of tracking members interacting with the item.

%% file: experiments.tex
\section{Experiments}\label{sec:experiments}
In this section we present variety of vertical applications to demonstrate how GNNs can be applied to production. We will show ablation studies and online A/B test impact.

\subsection{Experiments in Follow Feed}\label{sec:follow_feed_experiments}
\input{experiments_feed.tex}

\subsection{Experiments in Out-Of-Network Feed}\label{sec:oon_experiments}
\input{experiments_oon.tex}

\subsection{Experiments in Job Recommendations}\label{sec:Jobs_experiments}
\input{experiments_lts.tex}

\subsection{Experiments in People Recommendations}\label{sec:pymk_experiments}
\input{experiments_pymk.tex}

\subsection{Experiments in Ads}\label{sec:ads_experiments}
\input{experiments_ads.tex}

%% file: experiments_feed.tex

The LinkedIn Follow Feed recommends posts from a member's professional network. GNN embeddings are used in the Embedding-Based Retrieval (EBR) model of the Follow Feed recommendation system. These embeddings effectively capture the viewer's relationship with the post creator and interest in the post content. In the GNN model, the Follow Feed recommendation issue is treated as a link prediction task, determining the likelihood of a member interacting with a post. This model employs a SAGE encoder for generating member and post embeddings and a cosine decoder to calculate the similarity between these embeddings, predicting interaction probability. The cosine similarity thus indicates the relevance between member and post. In the EBR model, the ranking score for topK candidate selection combines post recency and relevance (cosine similarity). An offline experiment showed the GNN-based EBR model achieves a relative 9.6\% improvement in recall compared to the existing rule-based candidate selection model. In the online A/B test we observe an relative improvement of 0.5\% in Feed Engaged Daily Active Users.

The ablation study for the Follow Feed GNN model offers insights for how to construct the GNN model to achieve the best model performance. Table \ref{table:follow-feed} highlights that including node ID embeddings significantly enhances model efficacy by an +15.3\% in validation AUC. The graph sampling strategy plays a crucial role, with performance generally improving as more neighbors are sampled; a jump from 20 to 200 neighbors results in an 3.2\% AUC increase. Different aggregators were evaluated, showing the attention aggregator  outperforms the mean and self-attention (where each node attends to itself) aggregators with an +0.9\% AUC increase. For link prediction tasks, using dual encoders (one for source and one for destination nodes) is more effective than a single encoder approach leading to an +2.5\% AUC. The sparsity of the Follow Feed graph shows that adding cold start edges, as outlined in (\S\ref{sec:coldstart}), leads to an 0.5\% improvement in validation AUC. The efficiency of sampling algorithms ranks as follows: 2-hop PPR Sampling > Random Sampling > Weighted Sampling. Implementing temporal modeling (\S\ref{sec:temporal_graphs}) in the Follow Feed application resulted in an 5.8\% AUC lift.


\begin{table}[]
\small 
\begin{tabular}{c|c}
\hline
Experiment Setup                & AUC Lift \\ \hline
Baseline: SAGE, 20 neighbors, Mean Aggregator & -   \\
SAGE, 200 neighbors & +3.2\% \\   
+ Attention Aggregator & +0.9\% \\
+ Dual Encoder & +2.5\% \\ 
+ ID embeddings & +15.3\% \\
+ Graph Densification & +0.5\% \\
 + 2-hop PPR Sampling & +0.6\% \\ 
 + Temporal Graph & +5.8\% \\
\hline
\end{tabular}
\caption{Follow Feed validation AUC for different GNN configurations. Techniques mentioned in the table are ordered in timeline of development and show incremental value of improvements on top of prior rows in the table.}
\label{table:follow-feed}
\vspace{-2.5em}
\end{table}

%% file: experiments_oon.tex

Beyond the in-network Follow Feed service, we've extended GNN applications to Out-Of-Network (OON) content recommendations. This feature allows LinkedIn members to discover content beyond their immediate network connections, tailored to their interests. OON posts are displayed on members' Feed pages or Notifications, based on engagement likelihood. We constructed the OON graph incorporating member-post engagement edges, member-creator affinity edges, and cold start edges (\S\ref{sec:Overview:GraphConstruction}, \S\ref{sec:coldstart}). The OON GNN model employs a SAGE-encoder and cosine decoder. The trained SAGE-encoder produces GNN embeddings for members and posts, which are then integrated into the Embedding-Based Retrieval (EBR) system as an additional candidate generator for OON recommendations. Our online evaluations indicated significant improvements in key metrics, including an approximate relative increase of 0.2\% in Daily Active Users (DAU) engaging with professional content.

%% file: experiments_lts.tex

GNN member embeddings are utilized in LinkedIn's Top Applicant Jobs (TAJ), a premium feature that suggests jobs to members with higher likelihood of being accepted. TAJ's ranking problem is framed as a link prediction task, using a heterogeneous graph with nodes representing members, jobs, skills, and positions (company-title pairs). The graph, enriched by diverse edge types, and connects members to jobs based on their application history. The GNN model, trained on this rich graph, produces member embeddings for integration into TAJ's ranking models. GNN embeddings which are added on top of the existing two-tower models (member-job) have shown substantial improvements in key metrics, both offline and online, as detailed in Table ~\ref{table:jobs_recommendations_table}.
Notable relative achievements include a 0.3\% increase in premium member subscription renewal, a 1\% rise in the hearing back rate (applications receiving a positive response within 7 days), and a 1.8\% growth in company follows. Additionally, temporal model in job recommendations yielded a 6.8\% AUC lift in the Job Recommendation Ranking model, leading to relative increases of 0.4\% in job viewers and 0.4\% in total qualified applicants.

\begin{table}[ht]
\centering
\small 
\begin{tabular}{l|cc}
\hline
\textbf{Evaluation} & Metric & Lift  \\ \hline
Offline & AUC & $+1.1\%$  \\ \hline
Online & Renewal Rate & $+0.3\%$ \\  
& Positive Hearing Back Rate& $+1.0\%$ \\
& Company Follows & $+1.8\%$ \\ \hline
\end{tabular}
\caption{Job Recommendation offline and online relative improvements. }
\label{table:jobs_recommendations_table}
\vspace{-2.0em}
\end{table}

%% file: experiments_pymk.tex
LinkedIn's people recommendation service suggests potential connections to members. GNN embeddings have been integrated into its retrieval phase. To train the GNN model, a substantial graph was constructed, comprising up to one billion member nodes and billions of connection edges. The weight of each edge between members $u$ and $v$ is determined by the formula:
$$\frac{\#\text{ of common connections between }u\text{ and }v}{\sqrt{\# \text{ of } u\text{'s connections}} \times \sqrt{\# \text{ of } v\text{'s connections}}}.$$
This use case has been formulated as a link prediction task. We compared 3 different multi-hop sampling strategies in Table \ref{table:pymk_sampling}. PPR family strategies outperform weighted sampling by at least 2\%. It is expected that multi-hop PPR sampling is slightly better than 2-hop PPR sampling but the improvement margin is relatively small. 
\begin{table}[H]
\begin{tabular}{c|c}
\hline
Sampling Method         & GNN Validation AUC Lift \\ \hline
2-hop Weighted Sampling & -                              \\ 
Multi-Hop PPR Sampling  & +2.3\%                        \\ 
2-hop PPR Sampling       & +2.1\%                        \\ \hline
\end{tabular}
\caption{GNN sampling in People Recommendations}
\label{table:pymk_sampling}
\end{table}
We choose 2-hop PPR sampling as the ultimate production sampling strategy due to the balance between metric enhancements and computational complexity. In particular, 2-hop PPR sampling contributes around 90\% of gains and accelerate the sampling speed by 3 times.
500M members' embeddings are generated and they are integrated as new features in downstream models for Embedding-Based-Retrieval (EBR) tasks. A summary of the offline recall and online metrics is presented in Table \ref{table:pymk_ebr}. We observe significant offline and online metric lift after incorporating GNN embeddings with 2-hop PPR sampling. 
\begin{table}
\begin{tabular}{c|cc}
\toprule
Evaluation              & Metric                & Lift    \\ \hline
Offline                 & Recall                 & +29.6\% \\ \hline
\multirow{3}{*}{Online} & Sessions               & +0.2\% \\ 
                        & Weekly Active User     & +0.1\% \\ 
                        & New Member Connections & +2.4\% \\ \hline
\end{tabular}
\caption{Relative improvements in online A/B experiments within People Recommendation EBR with GNN embeddings}
\label{table:pymk_ebr}
\vspace{-2.5em}
\end{table}

%% file: experiments_ads.tex
CTR prediction forms the cornerstone of LinkedIn's ads recommendation system, where GNN models are used to integrate graph topology into the prediction process. The ads graph includes member nodes, creative (ads) nodes, campaign nodes, and company nodes, connected by member interaction and creative attribute edges. However, the sparsity of the ads graph presents challenges. To mitigate this, we assumed members with similar tastes in Feed posts or similar connections might also share ad interaction patterns. Therefore, we added Feed affinity edges and member connection edges (with downsampling) to the graph, expanding it to billions of nodes and edges. In the post-GNN model training stage, we generated node embeddings and incorporated them into downstream CTR models. Offline metrics, summarized in Table \ref{table:ads_ctr}, show the impact of different edge types and GNN embeddings on CTR prediction. 
The results indicate that adding member GNN embeddings alone improves AUC by 0.17\%. The inclusion of Feed edges and member connection edges further increases AUC by 0.29\%. The most effective CTR model, showing an 0.39\% AUC lift, used all types of GNN embeddings, underscoring the synergy between member and item embeddings. When implemented online, this comprehensive CTR model, with member, creative, and campaign embeddings, achieved an approximate relative 2\% online CTR lift.
We are exploring boosting model performance without using Feed affinity and member connection. Graph densification becomes especially useful in this scenario for Ads. After adding artificial member-to-member edges via graph densification, we have seen an 0.28\% AUC lift. Billions of cross-domain edges can be safely replaced with adding artificial edges between members.

\begin{table}[H]
\begin{tabular}{c|cc}
\hline
Edges                                                                       & Output GNN Embeddings & AUC lift \\ \hline
Ads                                                                              & member  
& +0.17\%                      \\ \hline
\begin{tabular}[c]{@{}c@{}}Ads with \\ graph densification \end{tabular}  & member                            & +0.28\%                      \\ \hline
\begin{tabular}[c]{@{}c@{}}Ads, Feed affinity, \\ member connection\end{tabular} & member                            & +0.29\%                      \\ \hline
\begin{tabular}[c]{@{}c@{}}Ads, Feed affinity, \\ member connection\end{tabular} & \begin{tabular}[c]{@{}c@{}}member, \\ creative, campaign\end{tabular}     & +0.39\%                      \\ \hline
\end{tabular}
\caption{Offline Metrics for Ads CTR Models with GNN embeddings}
\label{table:ads_ctr}
\vspace{-2.0em}
\end{table}

%% file: deployment_lessons.tex
Over the development of GNNs at LinkedIn we experimented with variety of applications and training infrastructures. Here we share some of our deployment lessons we learnt during the development.

\subsection{Impression Discount Before Retrieval}
In Follow Feed experiments, even though models exhibit high metric improvement in offline assessment. However, when we deployed these models online, the initial positive effects fade away and sometimes turned negative after a few days. As this behavior is model agnostic, we examined the system-wise reasons for this behavior. We discovered that the impression discount component, which filters out the updates that members have already viewed, is located after the retrieval layer. Under this setting, the impression discount component will continuously discard the relevant updates chosen from the retrieval layer, since the relevance scores are relatively stable. The previous retrieval model does not face this issue, as it retrieves the most recent updates, which are dynamic. Online metrics improve steadily after we position the impression discount component before the retrieval model.

\subsection{Graph Engine scales up GNN training}
GNN models rely on sampled neighbors, or the "compute graph," for training. Initially, Spark jobs were used to pre-compute these graphs and store them on HDFS, but this approach had significant drawbacks. First, the precomputation process, especially self-joins on large graphs, was slow, taking 20 hours for about 500M nodes, and couldn't scale for billions of nodes. Second, training speeds were hampered by heavy disk I/O, as the precomputed graph data on HDFS was over 10 times larger than the original graph due to repeated nodes in the label data. Third, model iteration was sluggish since any change in neighbor sampling strategy required regenerating the compute graph. Additionally, the static nature of these precomputed graphs limited model generalization, affecting performance. To overcome these issues, we switched to using a Graph Engine for real-time compute graph sampling. This change eliminated the need for precomputation and addressed slow disk I/O by directly serving graph data from memory. With the graph data in GE, we can experiment with various sampling strategies or model architectures without altering the underlying graph, enhancing model iteration by 10X. Moreover, GE allows training jobs to dynamically request compute graphs in real-time for each training instance. Introducing randomness in sampling means compute graphs for the same node vary with each request, leading to better model performance through enhanced generalization.

%% file: conclusion.tex
In this paper we presented {\systemname}, large scale GNN framework at LinkedIn. 
We shared set of approaches to train GNN model effectively reducing training speed by 7x, and improving quality of baseline GNN model by large margin. The lessons we share in the paper can be useful to industry practitioners. {\systemname} has been deployed to variety of applications at LinkedIn including Feed, Jobs, people recommendation and Ads domains with significant production impact.

%% file: appendix.tex
\appendix
\section{INFORMATION FOR REPRODUCIBILITY}

\subsection{Parameter selection for training speed up using Adaptive Neighbor Sampling and Grouping \& Slicing}
\begin{figure} [h]
    \centering
    \includegraphics[width=1.0\columnwidth]{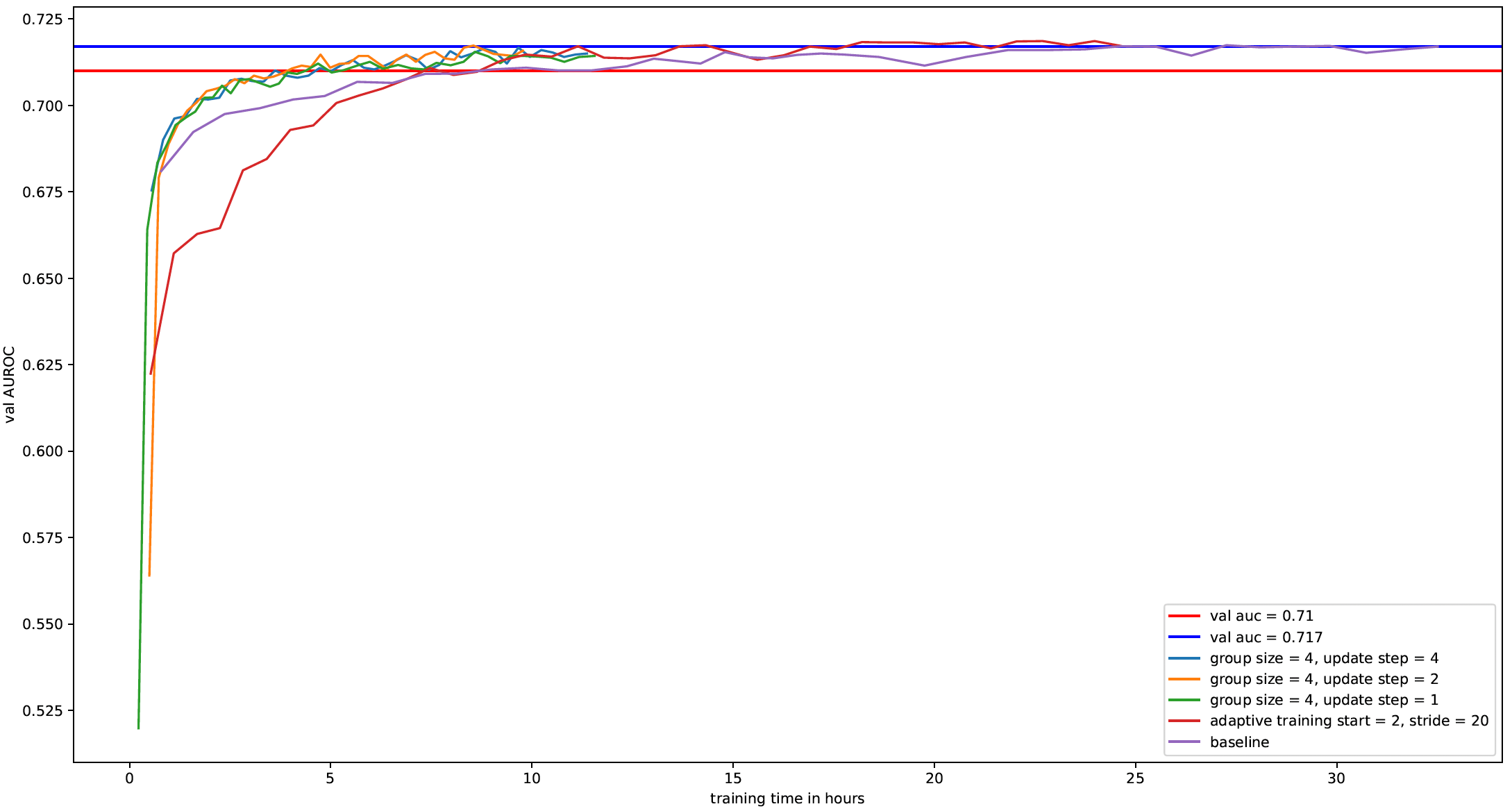}
    \caption{Follow Feed with Adaptive Neighbor Sampling and Grouping \& Slicing}
    \label{fig:feed-adp-grp-slc}
\end{figure}

\begin{figure} [h]
    \centering    \includegraphics[width=1.0\columnwidth]{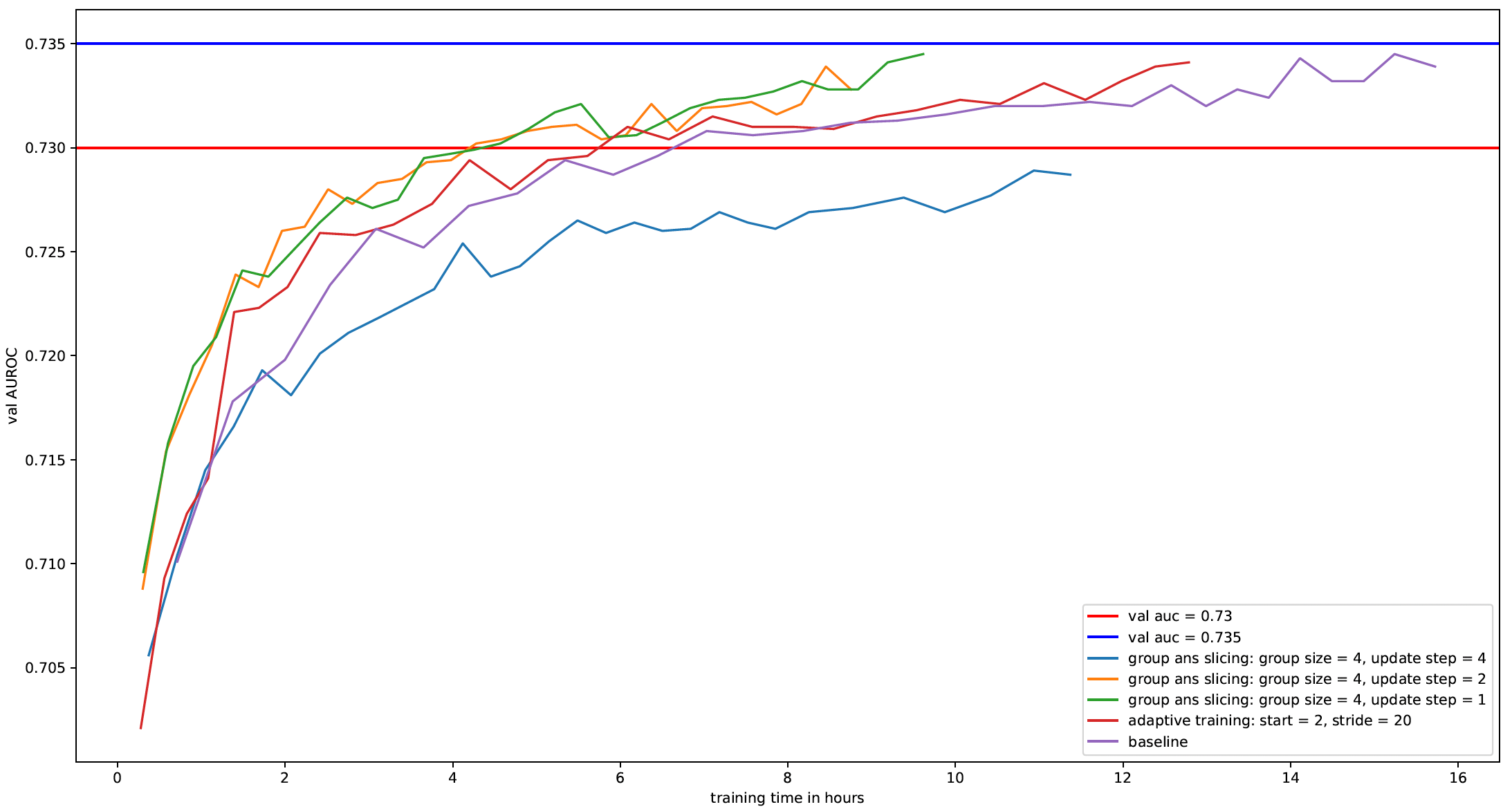}
    \caption{Job recommendation with Adaptive Neighbor Sampling and Grouping \& Slicing}
    \label{fig:lts:lts-adp-grp-slc}
\end{figure}

We provide additional information on parameters of Adaptive Neighbor Sampling and Grouping \& Slicing strategies. From convergence plots on Figure \ref{fig:feed-adp-grp-slc} and Figure \ref{fig:lts:lts-adp-grp-slc} we observe significant speed up of training due to Adaptive Neighbor Sampling and Grouping \& Slicing techniques described in \S\ref{sec:training_speed}. We can see that convergence speed is improved in comparison to baseline. Adaptive Neighbor Sampling  and Grouping \& Slicing require parameters tuning to observe improvement. Here we set the parameter to group size equal 4 and update step equal 1 for the Follow Feed case, and same parameters perform well in Job Recommendations. For Adaptive neighbor sampling we start with neighborhood size of 2 neighbors and increase the number of neighbors sampled in stride of 20. Same parameters settings were used across applications for Adaptive neighbor sampling.


\subsection{Nearline inference for GNNs}
\begin{figure} [h]
    \centering
    \includegraphics[width=1.15\columnwidth]{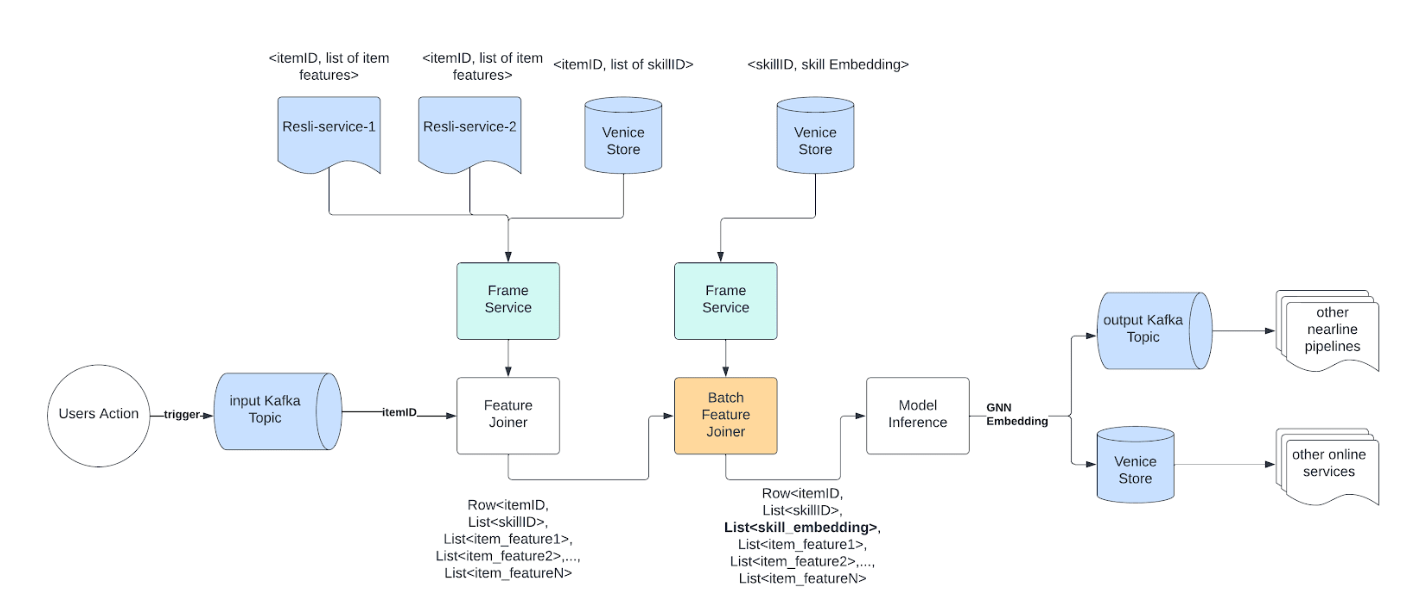}
    \caption{GNN nearline pipeline example}
    \label{fig:GNN_nearline}
\end{figure}

Here on Figure \ref{fig:GNN_nearline} we present details on nearline pipeline developed for Job Recommendations Engine. Due to deployment of the pipeline we could enable fresh job recommendations within LinkedIn.
